\documentclass{article}
\usepackage{spconf,amsmath}
\usepackage{amssymb}
\usepackage{amsfonts}
\usepackage[pdftex]{graphicx}
\usepackage{bm}
\usepackage{comment}

\newcommand{\ie}{i.e.}

\title{SCAPEGOAT GENERATION FOR PRIVACY PROTECTION FROM DEEPFAKE}
\name{Gido Kato$^{\star}$,~Yoshihiro Fukuhara$^{\star}$,~Mariko Isogawa$^{\ast}$ \\ \textit{Hideki Tsunashima$^{\star}$,~Hirokatsu Kataoka$^{\dagger}$,~Shigeo Morishima$^{\ddag}$} \thanks{This research is supported by the JST ACCEL Grant Number JPMJAC1602,  JST-Mirai Program Grant Number JPMJMI19B2, JSPS KAKENHI Grant Number JP21H05054 and JP19H04137.}}
  
\address{$^{\star}$ Waseda University \quad
    $^{\ast}$ Keio University~/~JST Presto \quad
    $^{\dagger}$AIST \\ 
    $^{\ddag}$Waseda Research Institute for Science and Engineering}
\begin{document}
%
\maketitle
\begin{abstract}

To protect privacy and prevent malicious use of deepfake, current studies propose methods that interfere with the generation process, such as detection and destruction approaches. However, these methods suffer from sub-optimal generalization performance to unseen models and add undesirable noise to the original image. To address these problems, we propose a new problem formulation for deepfake prevention: generating a ``scapegoat image'' by modifying the style of the original input in a way that is recognizable as an avatar by the user, but impossible to reconstruct the real face. Even in the case of malicious deepfake, the privacy of the users is still protected. To achieve this, we introduce an optimization-based editing method that utilizes GAN inversion to discourage deepfake models from generating similar scapegoats. We validate the effectiveness of our proposed method through quantitative and user studies.

\end{abstract}
\section{Introduction}
Deepfake is a well-known technique for learning-based image and video synthesis, which has gained significant attention in recent years due to its ability to replace one person's face in an image or video with another. However, malicious use of deepfake technology can result in the creation of false statements or actions, posing a significant threat to individuals' privacy, portrait rights, and social credibility. 

To address this issue, researchers put forwards two main methods: deepfake detection and deepfake destruction. The former is used to identify fake images or videos. The latter disrupts the media synthesis process by using adversarial examples specifically designed to target the deepfake generation model. 
However, on the one hand, newer deepfake methods can evade detection. On the other hand, deepfake destruction techniques require improved generalization performance for unknown deepfake methods because the effectiveness of adversarial examples depends on the optimized model. Therefore, current approaches are inadequate in dealing with emerging deepfake methods.

\begin{figure}[t]
    \centering
    \includegraphics[width=1.0\linewidth]{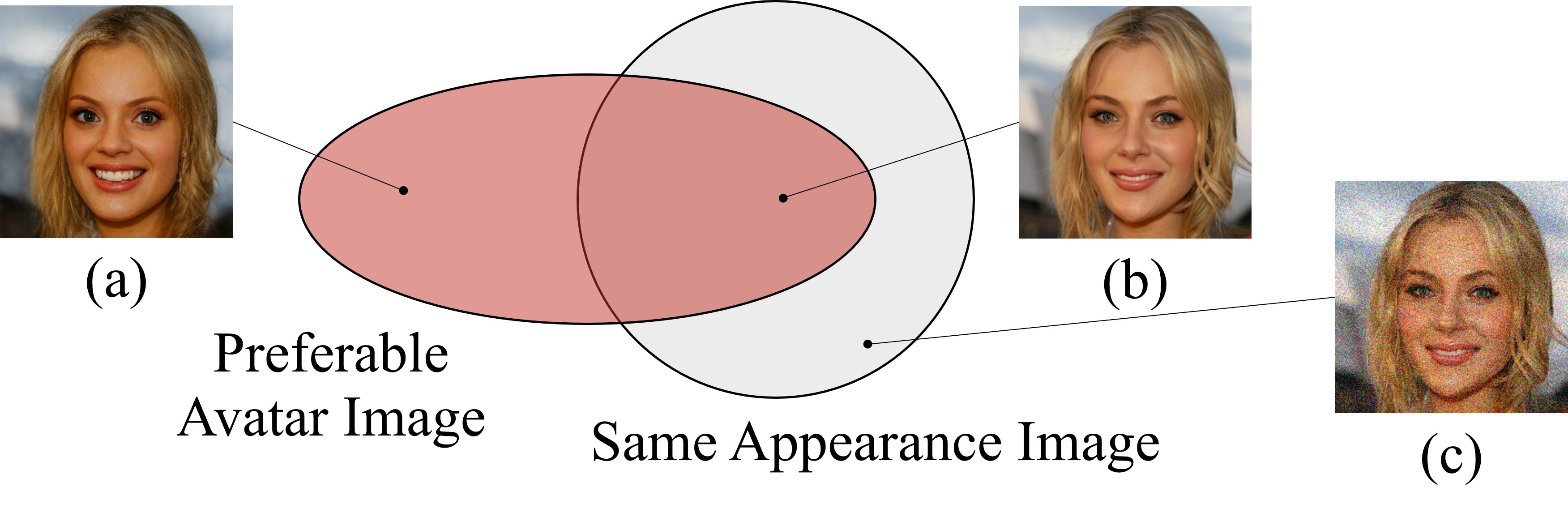}
    \vspace{-8mm}
    \caption{Overview of our problem set-up. (a) The proposed method of user-controlled editing will make the image look far from the original image but desirable as an avatar. (b) Original face image. (c) Existing privacy protection methods that add noise to the image may make the image look close but undesirable as an avatar.}
    \label{fig:issue}
\end{figure}

To address this limitation, we introduce an innovative direction called ``scapegoat generation'' that circumvents the generalization problem. Scapegoat generation modifies an image in a manner that is recognizable and preferred by the user, while ensuring it is impossible to restore the original input. This would effectively mitigate the privacy invasion risk, even in scenarios where deepfaked images have been extensively disseminated on the internet.
We propose a novel method in which users select specific facial features for editing and the editing is optimized to make the original face unguessable. 
We show our problem set-up overview in Fig.~\ref{fig:issue}, which represents images with the same appearance as the original image (gray) and images that users can recognize as their avatars (red).
The resulting image is referred to as a ``scapegoat image'', which appears distinct from the original image such that the original face cannot be reconstructed. Nevertheless, the user can still recognize the scapegoat image as their own. 
Our experiments evaluate the identity similarity between the generated scapegoat image and a deepfake of the scapegoat image against the original image, demonstrating that the scapegoat image effectively masks the facial features of the original image. 
Additionally, we evaluate facial similarity to the original image by user study, and the rating of the scapegoat image is significantly higher than that of the deepfake.
Our contributions are as follows: (1) introducing a novel problem setting called ``scapegoat generation'' that aims to mitigate privacy risks posed by deepfake: creating scapegoat images that users can recognize as their own and feel comfortable being deepfaked; (2) proposing a baseline method for creating such scapegoat images, and evaluate its effectiveness using both quantitative evaluation and user studies.

\section{Related work}

Deepfake is a deep learning-based media synthesis technology.
Recent research has led to the proposal of many face image editing techniques using deep learning. 
For instance,  
Simswap~\cite{chen2020simswap} leverages image ID information to swap the face of one person with another. While these techniques offer the ability to replace faces or modify expressions, they also create a potential for malicious use, such as creating defamatory images or videos that harm individuals.
To prevent deepfakes, two main approaches have been proposed: deepfake detection and deepfake destruction. The former aims to identify fake images by detecting artifacts unique to synthesized images. In recent studies, various techniques have been introduced to enhance detection performance~\cite{Shiohara_2022_CVPR, Liu_2021_CVPR, Haliassos_2021_CVPR}.
Deepfake destruction~\cite{ruiz2020disrupting,  sun2020landmark, yang2021defending, 10.1145/3503161.3547923}, on the other hand, refers to techniques that use more aggressive approaches than deepfake detection to prevent the generation of deepfakes.
The basic strategy of them is to add an adversarial perturbation~\cite{madry2018towards} to the input image against the deepfake model. 
As deepfake technology continues to evolve, it may become more difficult to detect and prevent deepfake using existing methods. Therefore, it is crucial to improve the generalization performance of both approaches.
Our proposed problem set-up is an innovative direction that avoids generalization challenges.

\section{Adversarial facial style modification}

\begin{figure*}[ht]
    \centering
    \includegraphics[width=0.8\linewidth]{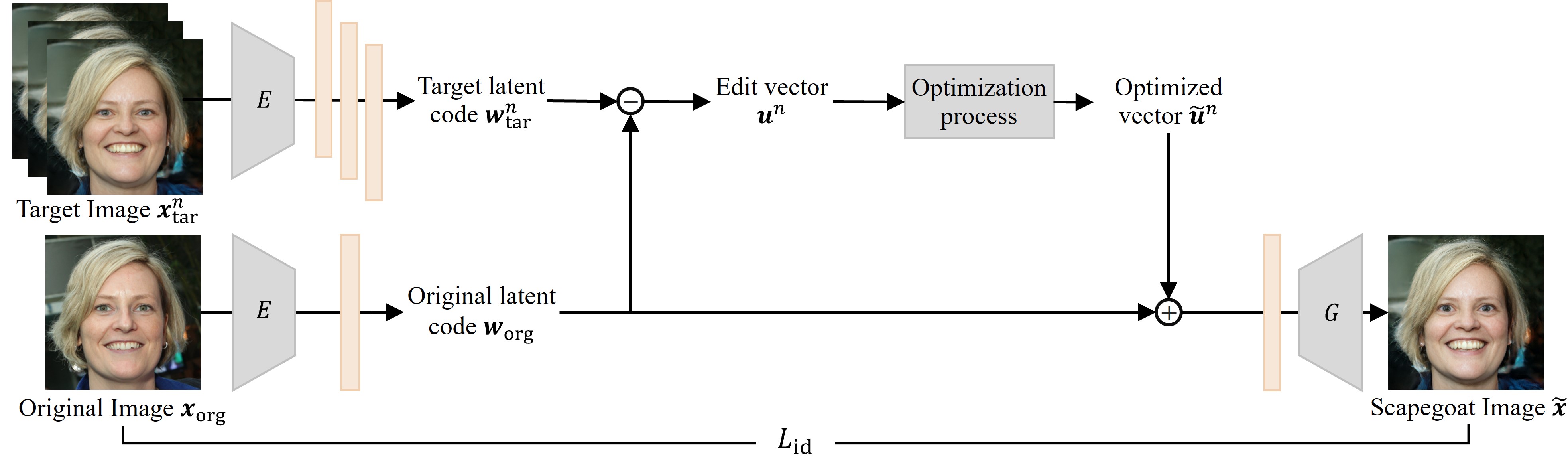}
    \vspace{-4mm}
    \caption{Overview of our proposed method. First, we embed the original image~$\bm{x}_{\mathrm{org}}$ and each target image~$\bm{x}_{\mathrm{tar}}^n$ for editing into the latent space of StyleGAN2~\cite{Karras_2020_CVPR}. Next, we compute the edit vector~$\bm{u}^n$ by taking the difference between the embedded original latent code~$\bm{w}_{\mathrm{org}}$ and the latent code of each target~$\bm{w}_{\mathrm{tar}}^n$. Then, we obtain the optimized vector~$\Tilde{\bm{u}}^n$ that produces the scapegoat~$\bm{\tilde{x}}$ image by optimizing each edit vector to maximize the loss~$\mathcal{L}_{\mathrm{id}}$. Finally, the scapegoat image~$\bm{\tilde{x}}$ is privacy-hidden to the extent specified by the user. Here we fix all parameters of the model.}
    \label{fig:proposed_method}
\end{figure*}
\begin{figure}[ht]
    \centering
    \includegraphics[width=1.0\linewidth]{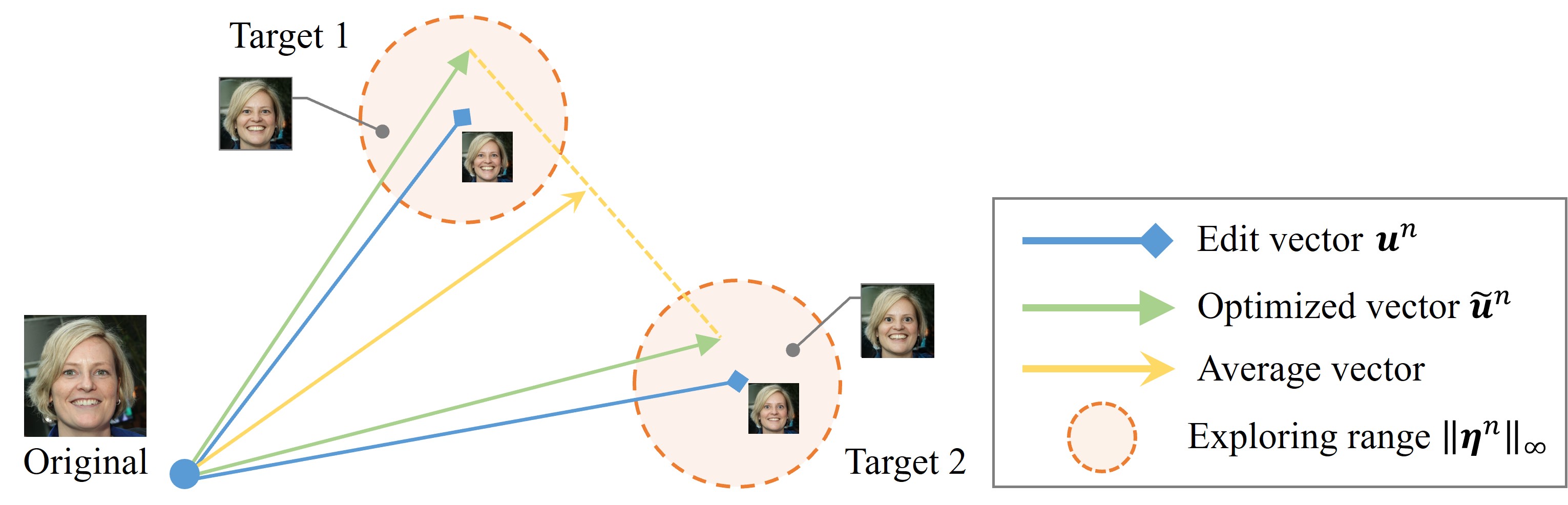}
    \vspace{-8mm}
    \caption{Overview of our optimization process. We search for the vector~$\bm{u}^n$ that maximizes the loss~$\mathcal{L}_{\mathrm{id}}$ in the neighborhood of each edit vector and average those vectors to obtain the final edit vector.}
    \label{fig:optimization_process}
\end{figure}
\subsection{Task overview}
First of all, we describe the existing problem set-up of deepfake destruction since our proposed method is based on the deepfake destruction method.
Existing deepfake methods input the original image $\bm{x}_{\mathrm{org}}$ into the deepfake generative model $DF$ to obtain a deepfake image $\bm{y}_{\mathrm{org}}=DF(\bm{x}_{\mathrm{org}})$.
Deepfake generally works to match the identity obtained using the face identification model $f$ with the input and output images, \ie , $f(\bm{x}_{\mathrm{org}})\simeq f(\bm{y})$.
Existing deepfake destruction methods interfere with deepfake generation by adding an optimized adversarial perturbation to the image, which makes the output of the deepfake generation model has a different identity than the original image. The optimization objective is $\max_{\bm{\eta}} [1-\cos( f(\bm{x}_{\mathrm{org}}), f(DF(\bm{x}_{\mathrm{org}}+\bm{\eta})))]$, where $\cos(\cdot,\cdot)$ is the cosine similarity.
However, in this problem setup, the performance of the adversarial perturbation depends on the optimized model and may not apply to unknown deepfake models.

In stark contrast, we propose a new problem setting called ``scapegoat generation'' to prevent the invasion of privacy by deepfake. 
We generate a scapegoat image $\tilde{\bm{x}}$ with editing that separates the identity of the image in advance, thereby hiding the privacy of the image. 
In our problem setup, while hiding all facial features in an image can prevent deepfakes, it may also lead to scapegoats that users cannot recognize as their own. 
To overcome this challenge, we suggest that users be given the ability to choose which facial features to hide and which to keep before editing their images. 
Our hypothesis is that this will allow for the creation of scapegoat images that maintain enough of the user's facial features to be recognized as an avatar while also concealing enough to prevent deepfake manipulation. 
The user defines the editing range by selecting target images and the editing process optimizes the image to ensure that the user's facial features are recognizable while still being immune to deepfakes.

\subsection{Our method}
We propose a method of scapegoat image generation in which the user specifies the extent of editing. The proposed method can generate scapegoat images that conceal privacy to the extent the user desires and can be deepfaked without any problems. To generate the scapegoat image, we employ an existing GAN inversion~\cite{Dinh_2022_CVPR} technique in which the image is embedded in the latent space of a trained GAN generator~\cite{Karras_2020_CVPR} and then semantically edited by adding a specific direction vector. Fig.~\ref{fig:proposed_method} shows a schematic diagram of the proposed method.
First, the user needs to prepare $N$ target images $\bm{X}_{\mathrm{tar}} = \left(\bm{x}_{\mathrm{tar}}^1, \bm{x}_{\mathrm{tar}}^2,\dots , \bm{x}_{\mathrm{tar}}^N\right)$ edited in advance in the direction they want (e.g. eye openness, smiling, face roundness), along with their face image $\bm{x}_{\mathrm{org}}\in \mathbb{R}^{H\times W\times 3}$ for editing. 
We assume that the edit axes specified by the user are those that the user is willing to hide with their facial features.
Then, we use the pre-trained encoder $E$ to embed the images into the latent space of the GAN to obtain the latent codes $\bm{w}_{\mathrm{org}}=E(\bm{x}_{\mathrm{org}}), ~\bm{w}_{\mathrm{tar}}^n=E(\bm{x}_{\mathrm{tar}}^n)$.
We take the difference between the embedded original latent code $\bm{w}_{\mathrm{org}}\in \mathbb{R}^{18 \times 512}$ and each target latent code $\bm{w}_{\mathrm{tar}}^n\in \mathbb{R}^{18 \times 512}$ to obtain the edit direction vector $\bm{u}^n=\bm{w}_{\mathrm{tar}}^n-\bm{w}_{\mathrm{org}}$ from the original image to the target image.
We optimize the edits in the neighborhood of each of these vectors that are the furthest away from the source image in terms of identity. Thus, we compute the final edit vector $\tilde{\bm{u}}^n$ by adding an optimized perturbation $\bm{\eta}^n$ to each of the edit vectors. We show a schematic diagram of the optimization in Fig.~\ref{fig:optimization_process}.
\begin{equation}
    \Tilde{\bm{u}}^n=\bm{u}^n+\bm{\eta}^n \quad s.t. \quad \|\bm{\eta}^n \|_{\infty} \leq \epsilon
\end{equation}

\noindent where, $\epsilon$ is the maximum value of the $L_{\infty}$ norm of the perturbation. The perturbation is optimized to maximize the loss function $\mathcal{L}_{\mathrm{id}}=1-\cos{(f(\bm{x}), f(\tilde{\bm{x}}))}$.
Then, the average of each optimized edit vector $\Tilde{\bm{u}}^n$ is added to the original latent code $\bm{w}_{\mathrm{org}}$ and input to the trained GAN generator $G$ to output the final scapegoat image $\tilde{\bm{x}}$.
\begin{equation}
    \tilde{\bm{x}}=G(\bm{w}_{\mathrm{org}}+\frac{1}{N}\sum_{n=1}^{N}\Tilde{\bm{u}}^n)
\end{equation}

The optimizer updates the perturbation in $k$ iterations. The equation for the $t$ th perturbation update is as follows:
\begin{align}
    \begin{split}
        \Tilde{\bm{u}}^n_t = \Tilde{\bm{u}}^n_{t-1} + a \ \mathrm{sign}[\nabla_{\Tilde{\bm{u}}^n_{t-1}}\mathcal{L}_{\mathrm{id}}] \\ 
        s.t. \quad \|\Tilde{\bm{u}}^n - \bm{u}^n\|_{\infty} \le \epsilon
    \end{split}
\end{align}
\noindent where, $a$ represents the step size of the update, and $a=0.01$ is used. 
The sign function returns 1, -1, or 0 depending on the sign of the number.

\section{Experiments}

\subsection{Experimental setup}
In our study, we employed StyleGAN2~\cite{Karras_2020_CVPR} pre-trained on the FFHQ~\cite{Karras_2019_CVPR} dataset as the GAN generator, and we utilized the HyperInverter~\cite{Dinh_2022_CVPR} to perform GAN inversion. 
We used Simswap~\cite{chen2020simswap} as deepfake models. 
Projected Gradient Descent(PGD)~\cite{madry2018towards} with $k=100$ iterations was utilized for optimizing scapegoat images.
The output ID loss was the cosine similarity of the ID vectors extracted by ArcFace~\cite{Deng_2019_CVPR}. 
We used the CelebA-HQ dataset~\cite{karras2018progressive, Liu_2015_ICCV} as our test set and generated scapegoat images. 
To generate the target images for the scapegoat generation, we used images that had been previously edited for three attributes (displeased, eye\_openness, face\_roundness) with HyperInverter. We denote the input of the deepfake method as $\bm{x}_{\mathrm{org}}$ (i.e., the source face) and $\bm{x}_{\mathrm{swap}}$ (i.e., the target face). These images were input into the deepfake generative model to obtain a deepfake image $\bm{y}_{\mathrm{org}}=DF(\bm{x}_{\mathrm{org}}, \bm{x}_{\mathrm{swap}})$. We randomly selected images from the test set to be used as $\bm{x}_
{\mathrm{swap}}$ in the deepfake face swap process.

\subsection{Results of scapegoat generation}

\begin{figure}[tb]
    \centering
    \includegraphics[width=1.0\linewidth]{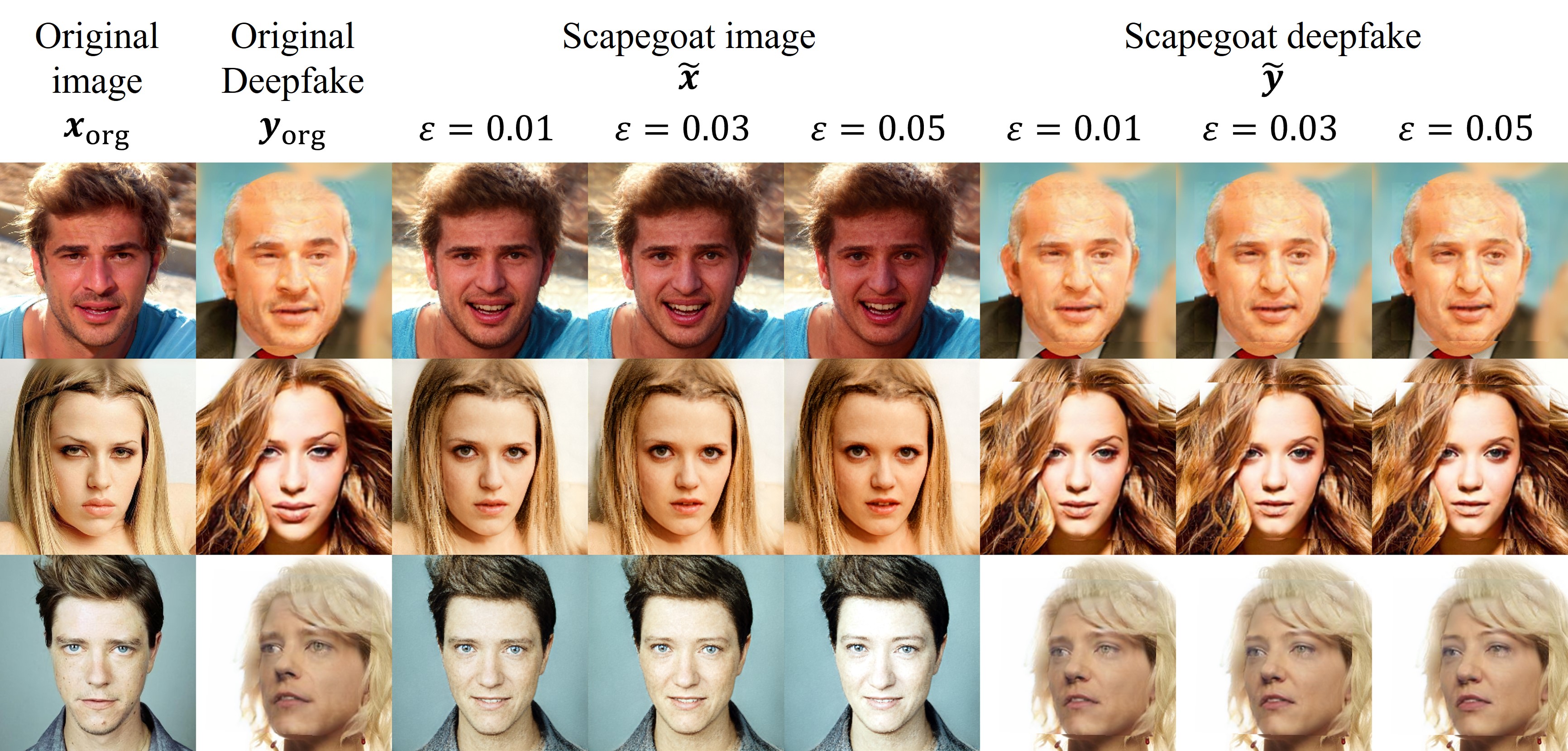}
    \vspace{-7mm}
    \caption{Results of scapegoat image generation. We used Simswap~\cite{chen2020simswap} as the deepfake model and CelebA-HQ~\cite{karras2018progressive, Liu_2015_ICCV} as the test set.}
    \label{fig:edit_eval}
\end{figure}
\begin{table}[!ht]
 \caption{Results of quantitative evaluation. We selected 3000 images from CelebA-HQ~\cite{karras2018progressive, Liu_2015_ICCV} to generate scapegoat images and evaluated their identity similarity to the original images. Note that we excluded samples for which face detection was impossible when performing deepfake.}
 \vspace{1mm}
 \label{table:quant_eval}
 \centering
  \begin{tabular}{ccc}
   \hline
   $\epsilon$ & $\mathcal{L}_{\mathrm{id}}$(edit) & $\mathcal{L}_{\mathrm{id}}$(deepfake) \\
   \hline \hline
   0.00 & - & 0.855($\pm$0.116) \\
   0.01 & 0.569($\pm$0.147) & 0.484($\pm$0.154)\\
   0.03 & 0.490($\pm$0.163) & 0.342($\pm$0.141)\\
   0.05 & 0.254($\pm$0.132) & 0.203($\pm$0.130)\\
   \hline
  \end{tabular}
\end{table}
To investigate the privacy protection capability of the proposed method, we evaluated the similarity of the generated scapegoat images and deepfaked images with respect to the original images.
A lower identity similarity between the generated images and the original image indicates a better privacy protection performance. 
We utilized a test set consisting of 3000 images randomly selected from the CelebA-HQ dataset. Qualitative results are presented in Fig.~\ref{fig:edit_eval}, and quantitative results are provided in Table~\ref{table:quant_eval}. 
The results in Table~\ref{table:quant_eval} demonstrate that the identity similarity decreases as the intensity of editing increases for both the scapegoat images and deepfake results. These outcomes suggest that the proposed method has a privacy protection effect.

\subsection{User study}

We conducted a user study to assess the quality of the scapegoat images. We started by creating a test set comprising 300 images randomly selected from the CelebA-HQ dataset~\cite{karras2018progressive, Liu_2015_ICCV}. We then generated eight versions of each image: the original image, an edited scapegoat image with three different intensity levels, and deepfakes of these 4 images using Simswap~\cite{chen2020simswap}. Subsequently, we distributed an image collection to each participant. The image collection consisted of 3 samples randomly selected from the 300 samples, without duplicates, arranged in a 3 x 8 table with eight images of one sample per row. Note that we arranged the generated images in a random order to prevent bias. During the experiment, we asked the participants to rate each image on a 7-point scale based on the original image to evaluate if it retained recognizable facial features. The evaluation criteria are as follows: 1) the image represents an entirely different person than the original image, 2) the image is almost a different person compared to the original image, 3) it is impossible to guess the original face, but there are similarities compared to the original image, 4) it is impossible to guess the original face, but the same person can be recognized compared to the original image, 5) the image retains enough of the person's character that the original face can be guessed, 6) the original face can be guessed from the image, and 7) the image represents the very person in the original image.
As our proposed method generates scapegoat images that users can comfortably recognize as their own image and feel comfortable with the potential for deepfake manipulation, we consider a rating of 3-4 on the 7-point scale to be appropriate for the scapegoat image.

\begin{figure}[!t]
    \centering
    \includegraphics[width=1.0\linewidth]{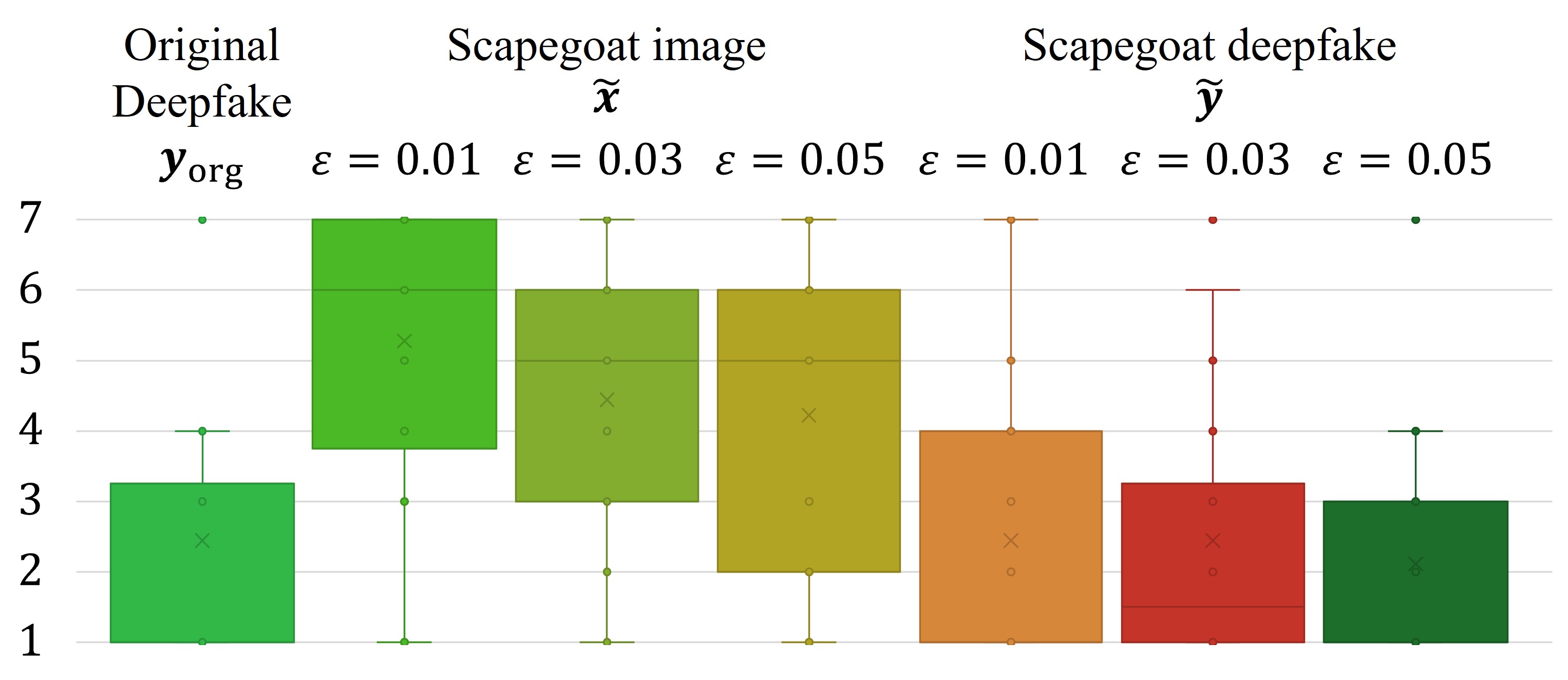}
    \vspace{-9mm}
    \caption{Boxplot of user study results. The boxes represent, from left to right, the deepfake results of the original image, the scapegoat image at each editing intensity, and the deepfake results of the scapegoat image.}
    \label{fig:boxplot}
\end{figure}
\begin{table}[!t]
 \caption{Table of p-values for a user study. ``*'' indicates a presence of significant difference.}
\vspace{1mm}
 \label{table:p-value}
 \centering
  \begin{tabular}{clll}
   \hline
   $\epsilon$ & $\bm{y}_{\mathrm{org}}-\Tilde{\bm{x}}$ & $\bm{y}_{\mathrm{org}}-\Tilde{\bm{y}}$ & $\Tilde{\bm{x}}-\Tilde{\bm{y}}$ \\
   \hline \hline
   0.01 & 0.0027* & 1.0000 & 0.0019* \\
   0.03 & 0.0137* & 1.0000 & 0.0057* \\
   0.05 & 0.0204* & 0.1696 & 0.0049* \\
   \hline
  \end{tabular}
\end{table}
We conducted a user study involving six participants and employed the Wilcoxon signed-rank test~\cite{10.2307/3001968} to analyze their responses using a significance level of $p < 0.05$. 
The results of our study are presented in Fig.~\ref{fig:boxplot}, and the corresponding p-values are reported in Table~\ref{table:p-value}. 
As shown in Fig.~\ref{fig:boxplot}: 
(1)	our proposed method is effective in processing the facial features of an image and concealing privacy, as evidenced by the decrease in evaluation scores with increasing editing intensity;
(2)	the deepfaked scapegoat image was generally evaluated as low, particularly when the edit intensity was $\epsilon=0.05$. Most evaluations fell within the range of 1-4 points, indicating that the subject was unable to recognize the face in the original image. Thus, we conclude that the deepfake scapegoat image does not invade privacy;
(3)	the significant variance observed in the evaluation values suggests that the degree of editing varied across images. Since we fixed the edit range in this study, users will need to interactively specify the edit range for each image in a future study.

\section{Conclusion}
In this paper, we present a novel problem setup called ``scapegoat generation'' and a baseline method for preventing privacy damage caused by deepfake by generating scapegoat images that users are willing to accept being deepfaked. Our proposed method utilizes GAN inversion techniques to optimize editing based on user-specified target images, resulting in privacy-preserving image generation while still recognizable as an avatar by the user. The effectiveness of the proposed method is confirmed through quantitative and user evaluation experiments. In future work, we aim to further improve our method by incorporating text-based editing, enabling the generation of scapegoats that better reflect the user's intentions.

\bibliographystyle{IEEE}

\begin{thebibliography}{10}

\bibitem{chen2020simswap}
Renwang Chen, Xuanhong Chen, Bingbing Ni, and Yanhao Ge,
\newblock ``Simswap: An efficient framework for high fidelity face swapping,''
\newblock in {\em Proceedings of the 28th ACM International Conference on
  Multimedia}, 2020, pp. 2003--2011.

\bibitem{Shiohara_2022_CVPR}
Kaede Shiohara and Toshihiko Yamasaki,
\newblock ``Detecting deepfakes with self-blended images,''
\newblock in {\em Proceedings of the IEEE/CVF Conference on Computer Vision and
  Pattern Recognition (CVPR)}, 2022, pp. 18720--18729.

\bibitem{Liu_2021_CVPR}
Honggu Liu, Xiaodan Li, Wenbo Zhou, Yuefeng Chen, Yuan He, Hui Xue, Weiming
  Zhang, and Nenghai Yu,
\newblock ``Spatial-phase shallow learning: Rethinking face forgery detection
  in frequency domain,''
\newblock in {\em Proceedings of the IEEE/CVF Conference on Computer Vision and
  Pattern Recognition (CVPR)}, 2021, pp. 772--781.

\bibitem{Haliassos_2021_CVPR}
Alexandros Haliassos, Konstantinos Vougioukas, Stavros Petridis, and Maja
  Pantic,
\newblock ``Lips don't lie: A generalisable and robust approach to face forgery
  detection,''
\newblock in {\em Proceedings of the IEEE/CVF Conference on Computer Vision and
  Pattern Recognition (CVPR)}, 2021, pp. 5039--5049.

\bibitem{ruiz2020disrupting}
Nataniel Ruiz, Sarah~Adel Bargal, and Stan Sclaroff,
\newblock ``Disrupting deepfakes: Adversarial attacks against conditional image
  translation networks and facial manipulation systems,''
\newblock 2020.

\bibitem{sun2020landmark}
Pu~Sun, Yuezun Li, Honggang Qi, and Siwei Lyu,
\newblock ``Landmark breaker: obstructing deepfake by disturbing landmark
  extraction,''
\newblock in {\em 2020 IEEE International Workshop on Information Forensics and
  Security (WIFS)}. IEEE, 2020, pp. 1--6.

\bibitem{yang2021defending}
Chaofei Yang, Leah Ding, Yiran Chen, and Hai Li,
\newblock ``Defending against gan-based deepfake attacks via
  transformation-aware adversarial faces,''
\newblock in {\em 2021 International Joint Conference on Neural Networks
  (IJCNN)}. IEEE, 2021, pp. 1--8.

\bibitem{10.1145/3503161.3547923}
Ziwen He, Wei Wang, Weinan Guan, Jing Dong, and Tieniu Tan,
\newblock ``Defeating deepfakes via adversarial visual reconstruction,''
\newblock in {\em Proceedings of the 30th ACM International Conference on
  Multimedia}, New York, NY, USA, 2022, MM '22, p. 2464–2472, Association for
  Computing Machinery.

\bibitem{madry2018towards}
Aleksander Madry, Aleksandar Makelov, Ludwig Schmidt, Dimitris Tsipras, and
  Adrian Vladu,
\newblock ``Towards deep learning models resistant to adversarial attacks,''
\newblock in {\em International Conference on Learning Representations}, 2018.

\bibitem{Karras_2020_CVPR}
Tero Karras, Samuli Laine, Miika Aittala, Janne Hellsten, Jaakko Lehtinen, and
  Timo Aila,
\newblock ``Analyzing and improving the image quality of stylegan,''
\newblock in {\em Proceedings of the IEEE/CVF Conference on Computer Vision and
  Pattern Recognition (CVPR)}, 2020, pp. 8110--8119.

\bibitem{Dinh_2022_CVPR}
Tan~M. Dinh, Anh~Tuan Tran, Rang Nguyen, and Binh-Son Hua,
\newblock ``Hyperinverter: Improving stylegan inversion via hypernetwork,''
\newblock in {\em Proceedings of the IEEE/CVF Conference on Computer Vision and
  Pattern Recognition (CVPR)}, 2022, pp. 11389--11398.

\bibitem{Karras_2019_CVPR}
Tero Karras, Samuli Laine, and Timo Aila,
\newblock ``A style-based generator architecture for generative adversarial
  networks,''
\newblock in {\em Proceedings of the IEEE/CVF Conference on Computer Vision and
  Pattern Recognition (CVPR)}, pp. 4401--4410.

\bibitem{Deng_2019_CVPR}
Jiankang Deng, Jia Guo, Niannan Xue, and Stefanos Zafeiriou,
\newblock ``Arcface: Additive angular margin loss for deep face recognition,''
\newblock in {\em Proceedings of the IEEE/CVF Conference on Computer Vision and
  Pattern Recognition (CVPR)}, 2019, pp. 4685--4694.

\bibitem{karras2018progressive}
Tero Karras, Timo Aila, Samuli Laine, and Jaakko Lehtinen,
\newblock ``Progressive growing of {GAN}s for improved quality, stability, and
  variation,''
\newblock in {\em International Conference on Learning Representations}, 2018.

\bibitem{Liu_2015_ICCV}
Ziwei Liu, Ping Luo, Xiaogang Wang, and Xiaoou Tang,
\newblock ``Deep learning face attributes in the wild,''
\newblock in {\em Proceedings of the IEEE International Conference on Computer
  Vision (ICCV)}, 2015, pp. 3730--3738.

\bibitem{10.2307/3001968}
Frank Wilcoxon,
\newblock ``Individual comparisons by ranking methods,''
\newblock {\em Biometrics Bulletin}, vol. 1, no. 6, pp. 80--83, 1945.

\end{thebibliography}

\end{document}